\documentclass[conference]{IEEEtran}
\IEEEoverridecommandlockouts

\usepackage{cite}
\usepackage{amsmath,amssymb,amsfonts}
\usepackage{algorithmic}
\usepackage{graphicx}
\usepackage{textcomp}
\usepackage{xcolor}
\usepackage{mdframed}
\usepackage{url}
\usepackage{makecell}
\def\BibTeX{{\rm B\kern-.05em{\sc i\kern-.025em b}\kern-.08em
    T\kern-.1667em\lower.7ex\hbox{E}\kern-.125emX}}
\begin{document}

\title{Social Group Human-Robot Interaction: A Scoping Review of Computational Challenges}

\author{\IEEEauthorblockN{Massimiliano Nigro}
\IEEEauthorblockA{
\textit{Politecnico di Milano}\\
Milan, Italy \\
massimiliano.nigro@polimi.it}
\and
\IEEEauthorblockN{Emmanuel Akinrintoyo}
\IEEEauthorblockA{
\textit{Imperial College}\\
London, United Kingdom \\
e.akinrintoyo23@imperial.ac.uk
}
\and
\IEEEauthorblockN{Nicole Salomons}
\IEEEauthorblockA{
\textit{Imperial College}\\
London, United Kingdom \\
n.salomons@imperial.ac.uk}
\and
\IEEEauthorblockN{Micol Spitale}
\IEEEauthorblockA{
\textit{Politecnico di Milano}\\
Milan, Italy \\
micol.spitale@polimi.it}}

\maketitle

\begin{abstract}
Group interactions are a natural part of our daily life, and as robots become more integrated into society, they must be able to socially interact with multiple people at the same time. However, group human-robot interaction (HRI) poses unique computational challenges often overlooked in the current HRI literature.
We conducted a scoping review including 44 group HRI papers from the last decade (2015-2024). From these papers, we extracted variables related to perception and behaviour generation challenges, as well as factors related to the environment, group, and robot capabilities that influence these challenges. Our findings show that key computational challenges in perception included detection of groups, engagement, and conversation information, while challenges in behaviour generation involved developing approaching and conversational behaviours.
We also identified research gaps, such as improving detection of subgroups and interpersonal relationships, and recommended future work in group HRI to help researchers address these computational challenges.
\end{abstract}


\section{Introduction}

Real-world public spaces, such as libraries, hospitals, and classrooms, are defined by the natural group interactions that occur among people. In these settings, robots have many opportunities to engage with people, but to interact successfully, they must be capable of engaging with multiple individuals at the same time \cite{chen2024integrating,cooper2023challenges,muller2023no,oliveira2021human,sebo2020robots,schneiders2022non}.  
Imagine a hospital where a robot at the reception helps a group of people—like a patient, their family, and a caregiver—all at the same time. Instead of focusing on one person, the robot listens to everyone's questions, figures out who is asking what, and answers each person clearly. For example, it might help the patient check-in, give directions to the family, and remind the caregiver about appointment times—all in one interaction. 
This group-focused interaction is more advanced than just talking to one person at a time.

These group settings introduce additional computational challenges compared to one-on-one interactions, which current robotic systems struggle to address. Most prior research in HRI has focused on dyadic interactions—between one robot and one person—leaving significant gaps in our understanding of how robots should operate in environments involving multiple people.
For example, robots may struggle in identifying group members as group dynamics can involve members moving in and out, blocking each other's views, and turning around \cite{muller2023no}.
Speech-to-text models also fail in transcribing speech in noisy environments with overlapping voices \cite{cooper2023challenges}, and tracking who is speaking and when remains an open challenge \cite{muller2023no}. 

Overall, group dynamics are inherently complex and continuously evolving. They are influenced by individual roles, interpersonal relationships, and group size, all of which can alter how people interact \cite{cartwright1968group}. These complexities create additional challenges for any robot attempting to engage in a group interaction.
But \textbf{what computational challenges do robots encounter in group interactions, and what research gaps need to be addressed to improve their ability socially interact effectively in group settings?} These are the key questions we aim to address in this survey. 

While several previous surveys have explored group interactions, they primarily focused on the interactional aspects of group dynamics—highlighting research trends, methodological challenges, or proposing frameworks to conceptualize group interactions. However, group robot interactions are difficult to develop. To fully leverage these insights and frameworks, significant computational advancements are required \cite{gillet2024interaction}.
This survey aims to bridge this gap by (1) describing the how and which computational challenge is currently being tackled in group HRI research, 
(2) identifying specific computational gaps, and (3) guiding future research in creating computational models that support robust group interactions.

Similar to Oliveira et al. \cite{oliveira2021human}, we consider group HRI to be any interaction between at least three group members who share a significant goal and who exert some type or degree of mutual influence over one another \cite{wilson1993groups}. To better identify challenges elicited by groups of people on a robot, we focus on groups of at least two people and one robot.

To guide this effort, we organize the current research on group robot interaction using the Input-Process-Output (IPO) framework adapted to group-robot interactions outlined by Sebo et al. \cite{sebo2020robots}.
The framework describes group interaction processes as influenced by ``input" factors, some examples being the setting of the interaction, size of the group and capabilities of the robot, and leading to output factors such as increased task performance for a group or increased social perception of the robot. We specifically focus on expanding the framework's group interaction processes section.
We also focus on understanding how \textbf{``input" factors} related to the group (e.g., group size), robot (e.g., robot sensing and acting capabilities), and environment (e.g., setting and context of the interaction) can shape and change the computational challenges. 
 We categorize the challenges involved in these processes into \textbf{perception} and \textbf{behaviour generation}. 

The following sections present the background on group HRI and its computational challenges, describe our methods for the scoping review, analyze the findings on perception and behaviour generation, and conclude with key takeaways, research gaps, and recommendations for future work.

\begin{figure*}
    \centering
    \includegraphics[width=\textwidth]{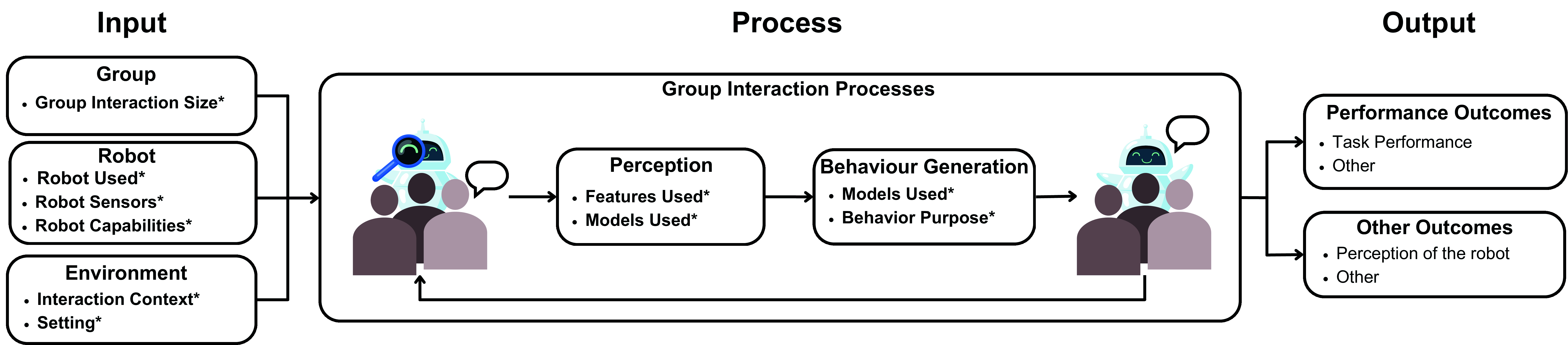}
    \caption{Extended IPO Framework illustrating perception and behaviour generation as part of group interaction processes.In bold and with asterisks the variables treated in this review. Note that this work does not take into account the output aspects of this framework as in \cite{sebo2020robots}.}
    \label{fig:block}
    
\end{figure*}

\section{Background}

\subsection{Previous Reviews on Group HRI}

Several previous works reviewed the state of group HRI, each highlighting different aspects. Oliveira et al. \cite{oliveira2021human} highlighted methodological issues in data collection methods for evaluating group HRI, such as questionnaires' limitations and the value of qualitative research methods. Furthermore, they highlighted transversal issues in group HRI research, such as the need for interdisciplinarity and reproducibility. 

Schneiders et al.\cite{schneiders2022non}  modified the 4C framework, an interaction framework focused on categorizing interactions between humans and digital artifacts, to categorize non-dyadic HRI. They used their framework to identify trends and highlight future directions on interaction modalities in non-dyadic HRI, which include using different types of tasks or contexts and the flow of the interaction to influence user experience. 

Gillet et al. \cite{gillet2024interaction} introduced Interaction-Shaping Robotics (ISR) as a subfield of HRI focused on robots that influence group behavior and attitudes. They identified key factors in these interactions and analyzed three human-robot group structures, each characterized by different compositions.

Weisswange et al. \cite{weisswange2024social} conducted a scoping review of social mediation of groups through robots. Using a modified IPO framework, which they termed the "Mediation IPO model," they developed 11 mediation approaches that robots can employ to shape group and team dynamics.

Sebo et al. \cite{sebo2020robots} instead focus on how robots influence groups, their future role in group interactions, and the steps researchers can take to understand better and implement group interactions. To this end, they extended the IPO framework for group interactions \cite{hackman1975group}. The framework places a group's interaction process between several input factors and outcomes.

Unlike previous reviews, this work focuses specifically on the computational aspects of group HRI.
According to Thomaz et al. \cite{thomaz2016computational}, computational HRI is defined as a subfield of HRI that emphasizes algorithmic and systems-oriented research. To explore these computational aspects, we adopt and expand upon Sebo's modified IPO framework. In our extended version, we conceptualize group interaction processes as a loop involving perception and behavior generation. In this way, the IPO framework also allows us to examine how inputs, such as group size or composition, influence the complexity of computational processes in group HRI.
\subsection{Perception}

Perception refers to how a robot processes and interprets sensor data, such as input from cameras, microphones, and other sensors, to gather valuable information about its internal and external environment \cite{stange2022self}. In group HRI, this involves interpreting sensor data to understand group dynamics and individual behaviours.
For instance, visual data can help the robot identify who belongs to the same group \cite{kollakidou2021enabling,schmuck2020robocentric,taylor2020robot,taylor2022group} and their positions relative to one another \cite{pathi2019estimating,barua2020let,kollakidou2021enabling,hedayati2020reform,hedayati2022predicting}, enabling the robot to determine the best way to approach and integrate into the group. Often, multiple data sources are combined: audio and visual data can work together to track who is saying what to whom \cite{matsusaka2003conversation}. Audio data allows the robot to understand the content of the conversation, while visual cues like gaze and head orientation help identify the addressee, enabling the robot to follow the conversation and recognize when it might take its turn.
More complex behaviours can also be analyzed through audio and visual inputs. For example, detecting when someone looks away or becomes silent can indicate disengagement \cite{leite2015comparing}. A tutoring robot, for instance, could use this information to identify which individuals are losing focus and take action to re-engage them \cite{leite2016autonomous}. Thus, perception allows the robot to understand the dynamics of a group interaction, which is fundamental to participate effectively in the interaction.

\subsection{Behaviour Generation}

Behaviour generation refers to creating and managing the execution of both communicative and non-communicative actions by the robot \cite{stange2022self}. In group HRI, the robot's behaviours are largely driven by its perception of the ongoing interaction. For example, once the robot understands how people are positioned within a group, it can determine the most appropriate and socially acceptable way to approach and join them \cite{yang2020impact,yang2019appgan,gao2019learning,samarakoon2018replicating,pathi2019estimating,kollakidou2021enabling,barua2020let}. Similarly, suppose the robot accurately grasps the flow and content of the discussion during a conversation. In that case, it can generate appropriate gaze behaviours \cite{tatarian2021robot,shintani2022expression,vazquez2017towards}, such as following the speaker or looking at an object being discussed \cite{vazquez2016maintaining}. Additionally, when the robot recognizes that someone is relinquishing their speaking turn, it can actively participate by deciding to take the turn. In this case, it might orient its body and gaze toward the person it intends to address and then produce a suitable response \cite{short2017robot,humblot2021talk}. In essence, behaviour generation builds on the robot's perception, enabling it to actively participate in group interactions by producing appropriate behaviours.

\section{Methodology}

We conducted a survey of HRI research that focuses on the main computational challenges in perception and behaviour generation that arise for robots when socially interacting with groups of people. To understand which computational challenge is unique to group interactions, we decided to make a comparison with dyadic interactions. Thus, we leveraged the PICO (Problem, Intervention, Comparison, and Outcome) framework, as outlined by Borrego et al. \cite{borrego2014systematic}:

\setlength{\leftmargini}{0pt} %
\begin{itemize}
    \item Problem: The scope of this research is group HRI studies with a focus on the technical design aspects.
    \item Intervention: The intervention considered in this work is the review of existing group HRI studies to identify current trends and practices.
    \item Comparison: A comparison is made with the well-explored dyadic interaction (interaction between one human and a robot) HRI studies to highlight key differences. 
    \item Outcome: The intended outcome of this research is to provide a technical guidebook for the design of the group HRI study and provide recommendations and identifying key research gaps to guide future work in group HRI.
\end{itemize}

We included papers published between 1st January 2015 and 1st July 2024. 
To maintain the focus of our review on studies that tackled computational challenges in social interaction we selected papers from the main HRI-specific venues. Thus similarly to Sebo et al. \cite{sebo2020robots}, our search was conducted on: (i) ACM/IEEE International Conference on Human-Robot Interaction (HRI), (ii) IEEE International Conference on Robot and Human Interactive Communication (RO-MAN), (iii) ACM Transactions on Human-Robotic Interaction (THRI), and (iv) International Journal of Social Robotics (IJSR).
 
The number of papers reviewed from each source was as follows: HRI (1239), RO-MAN (2,089), IJSR (860), and THRI (294), totalling 4,482 papers. Due to our cutoff date, we excluded papers from RO-MAN 2024 and the September issue of THRI, as these were released after late August 2024.

We defined the following inclusion criteria:
\begin{itemize}
    \item The work must involve at least one robot (whether physically present, simulated, or included in a dataset).
    \item Wizard of Oz (WoZ) studies are included only if the collected data is used within the same work to address a computational challenge in group HRI; otherwise, the robot must operate autonomously or semi-autonomously.
    \item The robot(s) must interact with at least two people.
    \item The robot must be socially interacting or about to initiate social interaction with people.
\end{itemize}

As specified in our inclusion criteria, we limited the inclusion of studies based on datasets or simulations unless they involved a robot. In this way, we ensured that the methods developed in these studies would be relevant and transferable to real-world robotic applications. For example, this is relevant to works focused on group detection. We excluded the study by Ramirez et al. \cite{ramirez2016modeling} because it relied solely on the Salsa dataset \cite{alameda2015salsa}, which consists of recordings from an overhead perspective of social gatherings, without applying or evaluating the system on a robot or a robot-recorded dataset. In contrast, we included the work from \cite{schmuck2020robocentric}, which tested their approach using a dataset collected by a robot in a real-world environment.
We excluded papers using the WoZ protocol for group interactions if they did not address a computational challenge in group HRI. For instance, Booth et al. \cite{booth2017piggybacking} investigated trust differences between groups and individuals in response to a teleoperated robot disguised as a food delivery service in university dormitories. However, since the robot was teleoperated, they did not tackle technical challenges such as how the robot identified groups or engaged in group conversations. In contrast, we included WoZ studies that leveraged the protocol to solve computational challenges in group HRI. An example is the aforementioned work from Schmuck et al. \cite{schmuck2020robocentric}, who used a WoZ protocol to collect data from a mobile robot in a real-world indoor environment and then used it to develop and evaluate a group detection system, offering valuable insights into this challenge.

Two researchers conducted the initial search and screening process. We screened the paper in two phases: first, we excluded papers that did not meet our inclusion criteria based on title and abstract, maintaining 79 papers, and then we conducted a full-text review, arriving at 44 full papers.

To highlight the key computational challenges in perception and behaviour generation for group HRI, we defined a set of variables to extract from each paper: 
the primary aspect of perception (e.g., group detection);
the primary behaviour generated (e.g., gaze);
the features (e.g., gaze direction) and models (e.g., clustering) used to implement perception;
 the models (e.g., generative models) used to generate behaviours and the purpose behind their generation (e.g., educating).

In addition, we extracted several variables that, while not directly related to the computational challenges themselves, impact both the complexity of these challenges and how researchers would solve them. We grouped these variables in: robot-related factors (robot used, robot sensors and capabilities); group-related factors (group size); environment-related factors (the study setting, interaction context).

In our expanded IPO framework (Figure \ref{fig:block}), we position the variables regarding perception and behaviour generation challenges as part of the group interaction processes. While robot, group, and environment-related factors are considered the ``inputs" to group interaction processes.
We chose not to elaborate on the ``outputs" of the IPO framework, as they do not directly impact the group interaction processes, which are the main focus of this review.

\section{Findings}

\subsection{Input} 

\begin{figure}[h] %
    \centering
    \includegraphics[width=0.3\textwidth]{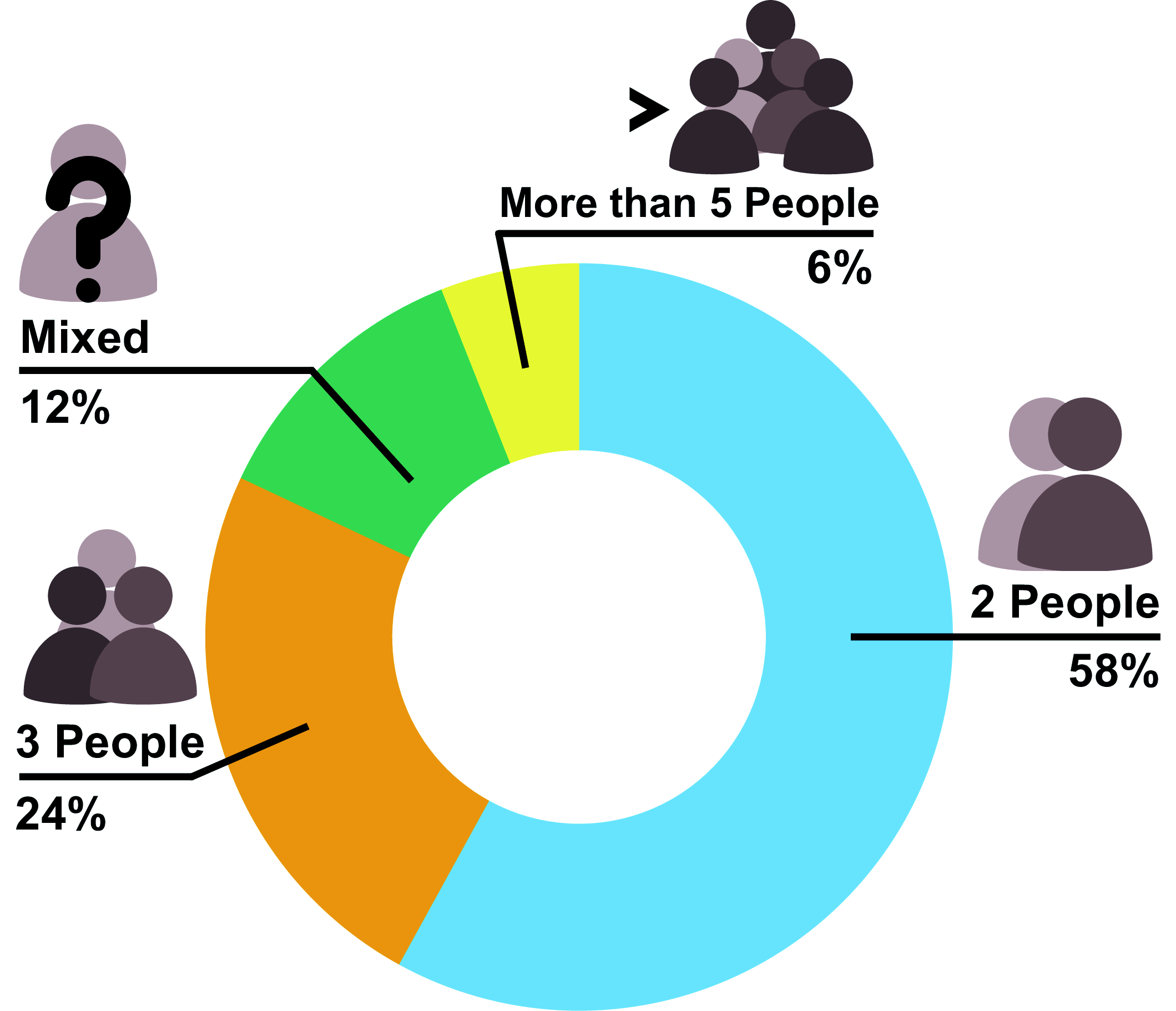} %
    \caption{Group size for studies involving a physical robot (robot excluded)}
    \label{fig:sample}

\end{figure}

In this section, we present the aggregated findings from our surveyed papers based on our selected input factors.

\subsubsection{Group}
One key factor that influences group dynamics and the computational challenges associated with it is the number of participants. As group size increases, so does the variability the robot must account for, complicating the interaction. As the group gets bigger, subgroups are more likely to form, people compete more for a chance to speak, and there is a greater chance of people talking over each other. Notably, 82\% of the studies we reviewed on autonomous robots focus on interactions with just two or three people, with 58\% focusing on two-person interactions.
 
\subsubsection{Robot}
Among the most commonly used robots, custom models led with 27\%, followed by Pepper, Nao, and Furhat, each at 12\%, and Emys at 9\%. Of the robots utilized, half are stationary and unable to navigate the environment. However, most can direct gaze (76\%) and speak (88\%), highlighting the importance of speech and gaze in social interaction.
Most robots rely on onboard sensors (59\%), with microphones being the most popular choice (59\%), followed by cameras (47\%) and RGBD cameras (12\%). To enhance perception, external sensors were also used, including Kinect (29\%), external cameras (12\%), and multiple microphones on people (32\%). 

\subsubsection{Environment}
The context of the interaction can also make perception or behaviour generation challenges less complex. For example, structuring a conversational interaction around a game or a problem-solving task can give cues about who will take the turn and what the conversation's argument is about. In our survey, 77\% of papers focused on interactions that were conversation-based, like games (34\%), tutoring (11\%), problem-solving tasks (8\%), team building activity (3\%),  but only 13\% were free unstructured conversations. 
Another 11\% of the papers robots executed service tasks while the remaining 13\% did not state the context of their work. 

Choosing a specific study environment can also limit the complexities of a group interaction. In 65\% of the reviewed studies with physical robots, experiments were conducted in controlled lab settings, with others being conducted in classrooms (18\%) and lower amounts in museums, community centers, shopping malls, and retirement homes. Lab-based studies often focused on conversational interactions, which become significantly more difficult in real-world environments due to unpredictable factors like noise, distractions, and overlapping speech. To overcome these challenges, many studies used additional external sensors to aid perception. For instance, in 25 studies focusing on conversational interactions, 11 employed external microphones for each participant to reduce issues related to noise and overlapping dialogue. Relying on extra microphones worn by individuals is impractical for a future where robots will be integrated into people's daily lives.

\begin{mdframed}
\textbf{Key Takeaway}: 
To reduce the complexities of computational challenges in group interactions, past works have limited group size to 2 or 3 people, structured conversations through games or tasks, conducted experiments in controlled lab environments, and used external sensors such as wearable microphones.
\end{mdframed}

\subsection{Perception}

\begin{table}[]
\centering
\begin{tabular}{|l|l|}
\hline
\multicolumn{2}{|c|}{\textbf{Perception}} \\ \hline
Engagement & \cite{zhang2021engagement,leite2016autonomous,nasir2022if,fan2021field,foster2017automatically,meng2020learning,leite2015comparing,alves2019empathic,tennent2019micbot} \\ \hline
Group and f-formation & \cite{barua2020let,schmuck2020robocentric,pathi2019estimating,taylor2020robot,hedayati2022predicting,taylor2022regroup,kollakidou2021enabling} \\ \hline
Conversation Information & \cite{shintani2022expression,tatarian2021robot,humblot2021talk,vazquez2016maintaining,tahir2020user,zarkowski2019multi,skantze2017predicting,neto2023robot,gillet2022learning,gillet2021robot,vazquez2017towards,utami2019collaborative} \\ \hline
Other & \cite{canevet2020mummer,brvsvcic2015escaping,gvirsman2024effect} \\ \hline
\end{tabular}
\vspace{1em}
\caption{Works included in each category for perception}
\end{table}

\begin{figure}[h] %
    \centering
    \includegraphics[width=0.48\textwidth]{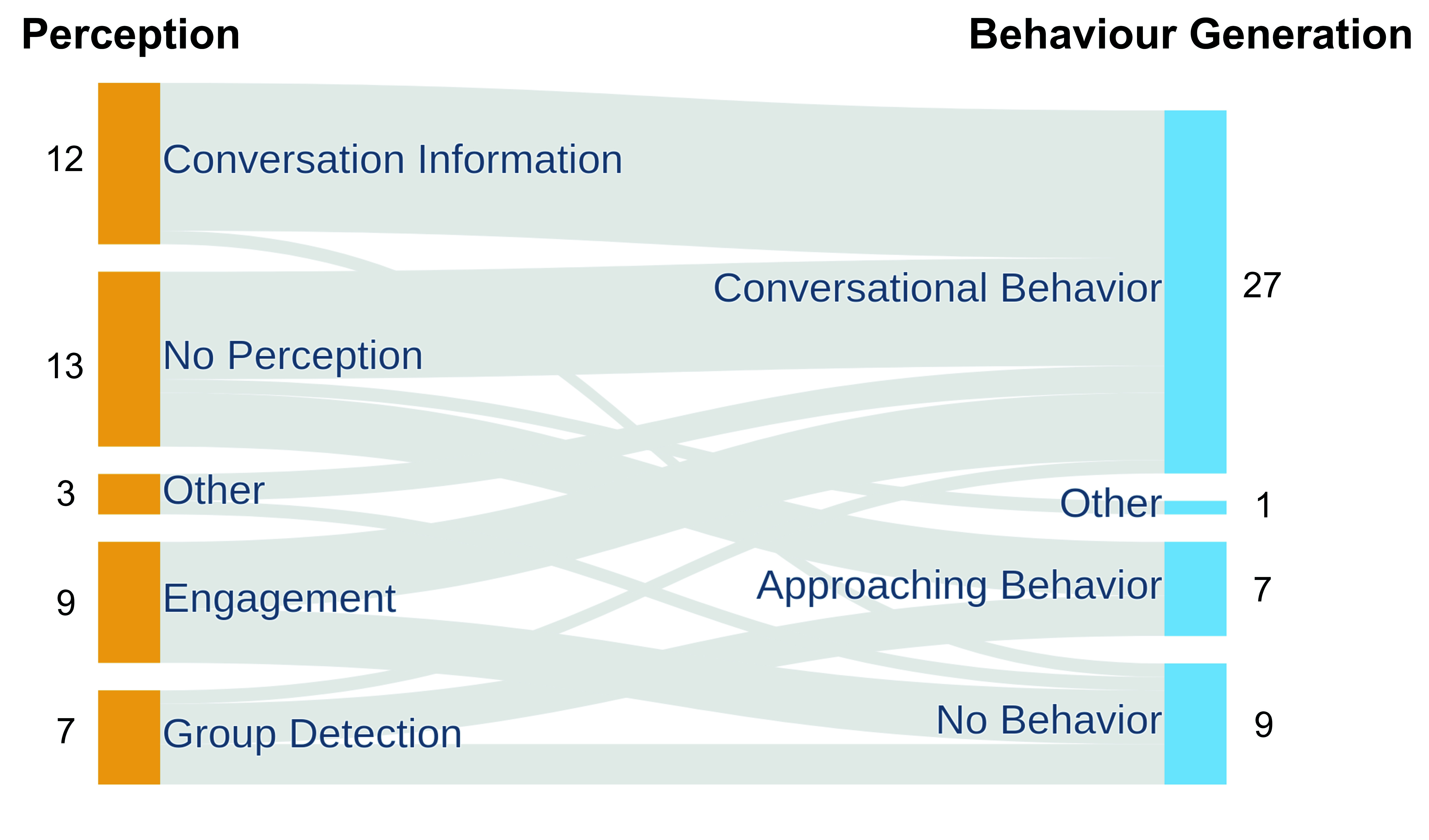} %
    \caption{Overview of key computational challenges in perception and behaviour generation and their interrelationships}
    \label{fig:overview}
    
\end{figure}

Our survey includes a total of 44 papers, of which 9 focused only on perception, 13 focused only on behaviour generation, and 22 focused on both as illustrated in Figure \ref{fig:overview}.

31 out of the 44 papers included in this survey worked on different challenges of perception for group robot interaction (\cite{zhang2021engagement,leite2016autonomous,nasir2022if} etc.). We categorized them into (i) group and f-formation detection (16\%), (ii) engagement detection (20\%), and (iii) detection of conversation information (27\%). We grouped the remaining challenges, like person re-identification and detection of robot abuse, under a general ``Other" category (7\%).

The following section will dive deeper into the key computational challenges we identified. We will expand on the differences between dyadic and group settings for these challenges and highlight the most commonly used features and models frequently employed to solve them in the literature.

\subsubsection{Group and F-Formation Detection} 

Group and f-formation detection systems are designed to identify which individuals belong to the same group and the spatial relationships they maintain as part of the group. A standard theory that describes the spatial relationships that groups adhere to is called f-formation (facing formation) \cite{kendon1990conducting}. Recognizing group membership and f-formations is essential for enabling robots to approach, join, and interact with groups in a socially acceptable manner. 
Significant progress in this area has been made, mainly due to the availability of public datasets \cite{alameda2015salsa, schmuck2020robocentric, alletto2014ego,Bazzani:CVPR12, hedayati2020reform}. 
Most papers in this category focus on detecting groups, with a minority also recognizing f-formation. 

The majority of works in our survey use people's position and orientation to identify either group \cite{ramirez2016modeling}, f-formation \cite{pathi2019estimating}, or both \cite{kollakidou2021enabling}.
Another used to detect groups is to combine the output of person detection models (e.g., YOLO \cite{7780460} and MobileNet-SSD \cite{younis2020real}) combined with depth information from the RGBD images. 
Most of the works used unsupervised learning approaches to perform group detection, with the majority using a variety of clustering approaches.
One approach worth highlighting is REGROUP \cite{taylor2022regroup}, which outperformed other state-of-the-art methods in group detection. REGROUP leverages person re-identification deep learning features to re-associate people to groups (github\footnote{\url{https://github.com/UCSD-RHC-Lab/regroup-hri}}). 

For f-formation classification, Pathi et al. \cite{pathi2019estimating} developed a set of geometric rules based on individuals' positions and orientations to detect different types of f-formations that are common in group settings such as l-shape, side-by-side, vis-a-vis or circular formation  \cite{kendon1990conducting}. 

\begin{mdframed}
\textbf{Key Takeaway:} Unsupervised learning approaches (e.g., clustering models), are effective for group detection starting from the output of person detection models.  
\end{mdframed}

\subsubsection{Engagement Detection}

Engagement detection has been one of the most extensively studied problems, with nine out of the 32 perception papers focusing on it. 
Following the definition given by Sidner et al. \cite{sidner2003engagement}, we refer to engagement as ``the process by which interactors start, maintain, and end their perceived connections to each other during an interaction".
In group interactions, people's behaviours are shaped by the roles they assume, the subgroups they belong to, and the strength of their relationships with others \cite{oertel2020engagement}, all of which can influence how they display disengagement. A clear example of this was found by Leite et al. They tested an engagement detection model trained on data collected from dyadic interactions with data collected from group interactions \cite{leite2015comparing}.
They discovered that participants who disengaged while interacting with their peers made the classification task more difficult. This is because peer interactions introduced a wider range of disengagement behaviours, requiring models to generalize to a much broader set of possible actions (such as talking with peers instead of simply looking away). 

Out of the nine papers in this category, two focus on detecting if a group of people would like to or is about to interact with a robot, namely estimating engagement intention \cite{zhang2021engagement,foster2017automatically}. For example, a robot bartender equipped with this capability could distinguish groups that want to order a drink from groups that are simply chatting at the bar \cite{foster2017automatically}. 
To estimate engagement intention, researchers frequently used features from whole-body posture and interpersonal distances, alongside gaze and head pose for each individual; with supervised classifiers like \cite{zhang2021engagement,foster2017automatically}.
The remaining seven papers \cite{leite2015comparing,leite2016autonomous,nasir2022if,fan2021field,meng2020learning,alves2019empathic,tennent2019micbot} instead focused on detecting if people were engaged during a social interaction. For example, understanding if a child is paying attention or is distracted by a robot-administered tutoring session. 
The most important features for assessing engagement during social interaction were audio (e.g., prosodies) or gaze features (e.g., children speaking with others or looking away may indicate they are not engaged in the tutoring session). 
Supervised classifiers, particularly Support Vector Machines as used by Leite et al. \cite{leite2015comparing}, have emerged as the most commonly employed method for engagement detection during social interactions.

\begin{mdframed}
\textbf{Key Takeaway:} When creating a model for engagement detection, the size of the group matters. Disengagement behaviours change with group size as people can disengage with the robot by interacting with each other.
\end{mdframed}

\subsubsection{Conversational Information Detection}
With the detection of conversation information, we refer to all the capabilities necessary for a robot to participate effectively in a group conversation. For example, detecting who and when a person is yielding or releasing a conversation turn \cite{zarkowski2019multi},people's roles in the conversation \cite{shintani2022expression,tatarian2021robot}, the group's focus of attention \cite{vazquez2016maintaining}, the social context of the conversation \cite{tahir2020user}, and participation equality between conversation members \cite{skantze2017predicting,gillet2021robot,gillet2022learning}.

In a one-on-one conversation, many things are implicit. Who is speaking, who is being addressed, and who is the next person who will speak is known by default. In group interactions, instead, these are complex computational problems \cite{gu2022says}. Additionally, people change their speaking and listening behaviour based on the number of participants in a conversation \cite{hadley2021conversation}, thus any approach aiming to understand emotional cues from these behaviours may face additional complexities in a group interaction compared to a dyadic one. 

Also, detecting conversation information in groups is particularly challenging in real-world settings, where background noise and overlapping speech frequently cause standard speech-to-text models, which are heavily relied upon, to perform poorly \cite{cooper2023challenges}. To address these challenges, most studies are conducted in controlled lab environments that allow the setup of multiple microphones so that every person's speech can be detected at any time with minimal noise. 

Researchers also use structured activities, such as games or problem-solving tasks where the robot assumes the role of a mediator, to better understand turn-taking behaviour by leveraging the contextual cues provided by these activities.
Even though this category groups solutions to many different problems, there are some common features in the context of a conversation.
Gaze direction was the most utilized feature, helping estimate dialogue roles \cite{tatarian2021robot}, focus of attention \cite{vazquez2016maintaining}, turn-taking \cite{zarkowski2019multi} and participation equality \cite{skantze2017predicting}. Users' proxemics, F-formation data, and group size were also used to determine dialogue roles \cite{tatarian2021robot}. Conversational features—e.g., who is speaking, when, how much—and the manner of speech were used to assess the conversation's social dynamics \cite{tahir2020user}.

Most of the selected studies used rule-based systems. Shintani et al. \cite{shintani2022expression} assigned dialogue roles like 'speaker,' 'addressee,' and 'side participant' based on turn-taking. Tatarian et al. \cite{tatarian2021robot} combined proxemics, gaze, and F-formation to classify the participants' roles as 'bystanders,' 'overhearers,' or 'active.' Vazquez et al. \cite{vazquez2016maintaining} used a rule-based approach to estimate the group's focus of attention via gaze features, while Zarkowski \cite{zarkowski2019multi} integrated gaze, microphone, and Kinect data to monitor the participant's turn-taking behaviour.
While rule-based systems can be effective in estimating dialogue roles and the focus of attention of a conversation, they are much harder to use to predict more abstract concepts like sociometrics or future conversation imbalance.
Thus, researchers utilized supervised learning techniques. 
Tahir et al. \cite{tahir2020user} compared machine learning models, finding SVMs most effective in classifying sociometrics like agreement, dominance, and interest, which were then mapped to different conversation's social contexts. Skantze et al. \cite{skantze2017predicting} utilized a multiple linear regression model to predict conversation imbalance by incorporating demographic data, prior imbalance, and gaze features.

\begin{mdframed}
\textbf{Key Takeaway}: Quick implementations like rule-based can still be useful for specific perception challenges in conversational settings, such as identifying dialogue roles or the group's focus of attention. 
\end{mdframed}

\subsubsection{Other}

Canevet et al. \cite{canevet2020mummer} tackled \textbf{person re-identification} by constructing a dataset of videos of people interacting with a robot in an open environment.
Then, using this dataset, they trained a classifier based on multiple detection modalities, such as face tracking and sound localization. 

Brscic et al. \cite{brvsvcic2015escaping} investigated the recognition of \textbf{robot abuse} episodes in a shopping mall setting. They gathered interaction data between people and robots in the mall to predict both the duration of interactions and the likelihood of abuse.

Gvirsman et al. \cite{gvirsman2024effect} incorporated \textbf{face detection and gaze tracking} into their robotic platform, Patricc \cite{gvirsman2020patricc}, enabling it to recognize users and direct its gaze toward them.

\subsection{Behaviour Generation}

\begin{table}[]
\centering
\begin{tabular}{|l|p{5cm}|}
\hline
\multicolumn{2}{|c|}{\textbf{Behavior Generation}} \\ \hline
Conversational Behavior & \makecell[l]{\cite{leite2016autonomous, tatarian2021robot, shintani2022expression,vazquez2017towards,vazquez2016maintaining,short2017robot,humblot2021talk,fan2021field,foster2017automatically,alves2019empathic,tennent2019micbot} \\
\cite{uchida2020improving,tahir2020user,zarkowski2019multi,strohkorb2020strategies,skantze2017predicting,kinoshita2017transgazer,brvsvcic2015escaping,gvirsman2024effect,neto2023robot,gillet2022learning,gillet2021robot,gvirsman2020patricc,oliveira2019stereotype,utami2019collaborative,correia2018group,strohkorb2018ripple}
} \\ \hline
Approaching Behavior & \makecell[l]{\cite{samarakoon2018replicating,pathi2019estimating,gao2019learning,yang2019appgan,yang2020impact,barua2020let,kollakidou2021enabling}} \\ \hline
Other & \cite{faria2021understanding} \\ \hline
\end{tabular}
\vspace{1em}
\caption{Works included in each category for behavior generation}
\end{table}

35 papers included in this survey worked on aspects relating to behaviour generation  for group robot interaction (\cite{shintani2022expression,tatarian2021robot,humblot2021talk} etc.). The generated behaviours are grouped into three categories which include (i) conversational behaviour (77\%), (ii) approaching behaviour (20\%), and (iii) other (3\%). 

\subsubsection{Conversational Behaviour}

With conversational behaviour, we refer to all the actions a robot must perform to participate effectively in a conversation. This includes turning toward the speaker, making eye contact, and directing speech to either an individual or the whole group.
In group interactions, deciding where to focus its gaze is more complex than in one-on-one conversations. With just one person, the robot only needs to focus on them. However, in a group, it has to decide whether to look at the speaker, the person being addressed or shift its gaze to an object if the conversation revolves around it, helping to establish a shared focus with the group.
Generating speech is also more challenging in group dynamics. In a one-on-one conversation, the robot only speaks to one person. In a group, however, it must decide whether to address a specific person, a subgroup, or everyone at once. When generating conversational behaviour, a robot must consider the group's culture, as it dramatically influences how people communicate. Some cultures prioritize harmony and consensus, while others open debate and personal expression, which leads to different ways of communicating\cite{feitosa2017influence}.

We classified twenty-five papers under the category of generating conversational behaviour. Most of these studies (65\%) focused on creating human-like and socially acceptable behaviour, enabling robots to engage in conversations in a more natural, human-like manner.
Meanwhile, 15\% of the works concentrated on developing repair strategies to re-engage users who may have become distracted. Another 10\% focused on balancing group interactions, ensuring that all participants contributed equally to the conversation. The final 10\% explored the use of speech and gaze to support educational interventions, to adjust the level of assistance provided, thereby making the learning experience more or less challenging.
88\% percent of the papers focused on generating conversational behaviour using rule-based approaches. These studies often explored the use of non-verbal cues, such as gaze \cite{kinoshita2017transgazer}, gestures, emotional behaviour \cite{correia2018group,strohkorb2018ripple} and subtle prompts \cite{oliveira2019stereotype} to influence group conversational dynamics \cite{gillet2021robot,gillet2022learning,tennent2019micbot,strohkorb2020strategies,neto2023robot}. For instance, robots can take on the role of implicit agents that encourage participation, as discussed in Tennent et al. \cite{tennent2019micbot}, drawing on Ju’s theory of implicit interactions \cite{ju2015design}. Alternatively, they may adopt a more active role by moderating discussions \cite{tahir2020user,utami2019collaborative,uchida2020improving} and directly addressing participants to ensure balanced participation \cite{short2017robot}.

Reinforcement learning (RL) was adopted in 8\% of the research. 
For example, Vazquez et al.~\cite{vazquez2016maintaining} used reinforcement learning to determine the orientation of a mobile robot during social group conversations. The robot was rewarded for turning towards the speaker in a conversation. However, RL-based research for behaviour generation requires more robust models and a lot of data (which is often missing in HRI studies) to capture the various human group dynamics, such as in conversations.
Another approach was used by Alves-Oliveira et al.~\cite{alves2019empathic}. Here, an autonomous robot was developed to teach students sustainable development. To learn the correct robot interaction strategies, they leveraged frequent pattern mining using data collected during a WoZ experiment. 

\begin{mdframed}
\textbf{Key Takeaway:} Datasets can be created effectively by using a WoZ approach to develop models that map perceived situations to specific actions, mimicking those in the collected data. 
\end{mdframed}

\subsubsection{Approaching Behaviour}
Approaching behaviour was investigated to enable robots to interact naturally and effectively with humans while maintaining social norms and expectations. Robots should approach humans without causing disruption, discomfort or violating their personal space. The behaviour of a robot should be predictable, comfortable and acceptable within the human space in which it operates~\cite{dautenhahn2006may, barua2020let, kollakidou2021enabling}. 

By adhering to social norms, robots will respect social conventions such as respecting interpersonal distances and knowing the correct angles that will be considered polite or non-intrusive to approach people from~\cite{dautenhahn2006may}. Such robots include service robots~\cite{avrunin2014socially} operating in hotels~\cite{nakanishi2018can}, restaurants~\cite{knight2024iterative}, airports~\cite{triebel2016spencer} and other public spaces~\cite{foster2019mummer}. 

The structure of the group and its dynamics must be identified to plan an approach that considers the boundaries and orientations of the persons. F-formation~\cite{kendon1990conducting} is mostly used by all research investigating how a robot should approach or join a group of people. Based on the perception assessment, a robot needs to understand the three spaces of the o-space (reserved for interaction), p-space (the space people stand in), and r-space (the area outside the interaction) to ensure it does not violate social norms. This is unlike dyadic interaction, in which a robot does not have to account for multiple social dynamics and spatial boundaries simultaneously. In fact,in group settings, the concept of group entitativity—how strongly individuals perceive their group as a cohesive unit—comes into play \cite{lickel2003elements}, as groups with higher entiativity will be more welcoming to the robot, thus preferring closer distances \cite{fraune2019human}. 

Seven papers belonged to the approaching behaviour category. Rule based method (57\%) was most commonly used for generating approaching behaviour. Machine learning (43\%) was another prevalent option. This comprises reinforcement learning (67\%) and generative models (33\%). Previous research explored various algorithms from simple techniques, such as by Kollakidou et al.~\cite{kollakidou2021enabling} which used a hierarchical clustering approach. While this was an efficient approach, it employs a custom distance function that considers basic position and orientation data for detecting group formations with discrete groups where there is a well-defined social grouping and predictable behaviour. Thus, it lacked the robustness for complex social scenarios. Thus, it may be unsuitable for real-world social settings unlike more sophisticated solutions such as those explored by Barua et al.~\cite{barua2020let} with machine learning algorithms. Barua et al. considered group detection and precise angle determination to handle outliers and occlusions better at the cost of higher computational demands. This highlights a trade-off between computational cost and real-time performance. This differs from dyadic interaction with lighter computation since a robot only has to maintain adequate personal space for a single person without having to find a position that is acceptable to every group member. 

Movement is essential for the study of approaching behaviour. Thus, mobile platforms (62.5\%) are the most commonly used for approaching behaviour studies. Movement was often combined with gestures, head gaze to interact with a group. Barua et al.~\cite{barua2020let} effectively combined multiple modalities such as skeletal key points, gestures, and movements to detect groups and determine an optimal approach. 

\begin{mdframed}
\textbf{Key Takeaway:} Designing systems for approaching behaviour requires balancing computational efficiency with precise angle determination, especially in complex social scenarios where the robot's view may be obstructed. Rule-based methods are efficient but often less accurate.

\end{mdframed}

\subsubsection{Other}

Faria et al. \cite{faria2021understanding} investigated how the \textbf{generated} behaviour of a robot with regard to \textbf{legible movements} can aid humans in interpreting its behaviour and intention. It achieved this by adopting a multi-user legibility model that calculates the trajectories of the robot and maximises the average legibility for multiple people with different perspectives.

\section{Research Gaps}
To identify research gaps, we analyzed psychology literature to uncover capabilities that could help robots interact with groups but are currently missing in group HRI research. Our analysis considered how human group interactions function and applied this knowledge to the computational challenges we identified in social group HRI. While the resulting list of capabilities is not exhaustive, we consider the identified gaps highly valuable for effective social group HRI.

\textbf{Detection of Subgroups.}
Our analysis suggests that none of the examined works focused on detecting subgroups but only aimed at detecting which people belonged to the same group and how the people were positioned in the group. In most instances, groups exhibit an internal structure composed of subgroups \cite{hogg2004social}. 
Members within these subgroups often collaborate closely during discussions and decision-making tasks \cite{hogg2004social}. Recognizing and managing subgroups becomes more and more important the larger the group. In fact, as group size increases, so does the potential for conflicts among subgroups and between a subgroup and the larger group. This complexity arises because interpersonal conflicts often escalate into conflicts between subgroups, impacting overall group dynamics significantly \cite{sidorenkov2020role}.
For robots that aim to facilitate interactions among people, recognizing the distinct subgroups is essential for effectively influencing these interactions.

\textbf{Detection of interpersonal relationships.}
Findings show that none of the papers we analyzed specifically addressed the detection of interpersonal relationships or incorporated interpersonal features. Understanding these relationships is essential, as they can significantly impact how individuals behave in group settings, offering valuable insights for perception tasks during conversations.
Salam et al. \cite{salam2016fully} demonstrated that integrating interpersonal features—such as the physical distance between participants and the attention they give and receive—enhanced model performance in classifying engagement within groups. This highlights the importance of including interpersonal features for perception tasks for group interactions.
Future research should investigate the effectiveness of interpersonal features in various perception tasks within group HRI. This could be particularly relevant for tasks like emotion recognition and the classification of social scenarios, where it is expected that individuals in a group could influence each other \cite{hess2022emotional}.

\textbf{Personalize approaching behaviours based on group entitativity}
Our findings show that none of the studies surveyed generates personalized approaching behaviours for groups. 
In dyadic HRI, numerous methods have been developed to personalize the approaching behaviours of robots to foster acceptance among individuals \cite{patompak2020learning,nigro2024interactive}. However, it is equally important to consider personalizing these behaviours to align with group preferences, as this could further enhance acceptance.
One significant factor that may influence accepted approaching behaviour is group entitativity; the more individuals perceive their group as cohesive, the more welcoming they are toward the robot, allowing for closer approaches \cite{fraune2019human}. Despite this, none of the methods reviewed for generating approaching behaviours in group interactions incorporated the personal preferences of groups. Future research should investigate whether tailoring approaching behaviours to group entitativity can increase robot acceptance in group settings.

\section{Recommendations \& Conclusion}

\textbf{Develop datasets and models for large groups.}
79\% of the studies involving autonomous robots focus on groups of two or three people, but larger groups may present additional challenges that still need to be studied. As group size increases, interaction dynamics can change significantly, with more subgroups forming and interpersonal relationships having a more significant impact on the interaction \cite{burke2006interactiton, delamater2006handbook}. 
For example, in Leite et. al, a group of three children disengaged by talking with each other, hence the model trained only on dyadic interactions was not effective in predicting these instances \cite{leite2015comparing}. We can imagine similar effects arise in similar perception problems, such as understanding group emotions or discerning between different social scenarios in a conversation. 
A potential solution is to gather datasets that include interactions across a range of group sizes, with a balanced amount of data for each group size. This could be achieved by designing interaction protocols that are flexible and scalable, making them easily applicable to groups of varying sizes. This solution, also helps identify which perception challenges are most affected by increasing group size.

\textbf{Train models for detection in group conversations with real-world data.}
Almost half of the studies focusing on conversational interactions with autonomous robots employed a microphone for each participant to reduce the impact of overlapping speech and background noise. While effective in controlled settings, this approach is not scalable to more complex, real-world environments where robots must rely solely on their own sensors. However, recent advancements in speech recognition show promise in improving a robot's ability to understand multiple speakers in noisy environments \cite{prabhavalkar2023end}. Future work should focus on collecting a dataset of group conversations between robots and people in loud environments. The dataset can then be used to fine-tune existing state-of-the-artmodels for speech detection or to train specific models for perception of group conversation. By training these models on more realistic, challenging scenarios, robots could become better equipped to handle the complexities of real-world human interactions without relying on external sensors.

\textbf{Take into account cultural factors when generating conversational behaviours.}
Eighty-eight percent of methods to generate conversational behaviour rely on simple rule-based systems. However, these one-size-fits-all models fail to account for cultural diversity. Culture significantly shapes communication styles and preferences, especially within groups \cite{feitosa2017influence}. These differences also affect decision-making and conflict resolution, with some cultures prioritizing harmony and consensus while others emphasizing open debate and self-expression. A robot using an inappropriate communication style could harm team cohesion and performance in collaborative settings. A promising approach is integrating cultural information into conversational models through embeddings. Future research can focus on creating datasets that reflect communication patterns across diverse cultures.

This work provides a comprehensive overview of the key computational challenges in group HRI, highlighting important insights from our analysis of each challenge. We also identified gaps in the current computational approaches to group HRI and proposed recommendations to advance the field.

However, we acknowledge that certain methodological choices in our review may have limited its scope. For example, other "input" factors that were out of the scope of the review (e.g., the robot's role) could also have an impact on the computational challenges in social group interactions. Similarly, relevant papers from more technically oriented robotics venues (e.g., IROS and ICRA) may have been missed due to our choice of focusing on HRI-specific sources.
Despite these limitations, we hope this work serves as a guide for researchers to overcome these challenges, ultimately advancing the deployment of robots in real-world group settings.

\bibliographystyle{IEEEtran}
\bibliography{IEEEabrv,bibliography}

\end{document}